\documentclass[12pt]{article} % For LaTeX2e

\usepackage[colorlinks=true]{hyperref}
\usepackage[margin=1.in]{geometry}
\usepackage{graphicx}
\usepackage{amsmath}
\usepackage{amssymb}
\usepackage{booktabs}
\usepackage{caption}
\usepackage{subcaption}
\usepackage{verbatim}
\usepackage{float}
\usepackage[english]{babel}
\usepackage{braket}
\usepackage{xy}

\title{Key Technology Considerations in Developing and Deploying Machine Learning Models in Clinical Radiology Practice}

\author{Viraj Kulkarni \\ \small{DeepTek Inc}
        \and Manish Gawali \\ \small{DeepTek Inc}
        \and Amit Kharat \\ \small{DeepTek Inc}}
        
\date{}

\begin{document}
\hyphenpenalty=1000

\maketitle

\begin{abstract}
\noindent The use of machine learning to develop intelligent software tools for interpretation of radiology images has gained widespread attention in recent years. The development, deployment, and eventual adoption of these models in clinical practice, however, remains fraught with challenges. In this paper, we propose a list of key considerations that machine learning researchers must recognize and address to make their models accurate, robust, and usable in practice. Namely, we discuss: insufficient training data, decentralized datasets, high cost of annotations, ambiguous ground truth, imbalance in class representation, asymmetric misclassification costs, relevant performance metrics, generalization of models to unseen datasets, model decay, adversarial attacks, explainability, fairness and bias, and clinical validation. We describe each consideration and identify techniques to address it. Although these techniques have been discussed in prior research literature, by freshly examining them in the context of medical imaging and compiling them in the form of a laundry list, we hope to make them more accessible to researchers, software developers, radiologists, and other stakeholders.
\end{abstract}

\section{Introduction}
Although radiology imaging has emerged as an indispensable tool in diagnostic medicine, there is a worldwide shortage of qualified radiologists to read, interpret, and report these images \cite{rimmer2017radiologist}\cite{nakajima2008radiologist}. The volume of images is growing faster than the number of radiologists. The high workload this causes leads to errors in diagnosis due to human fatigue, unacceptable delays in reporting, and stress and burnouts in radiologists. On the other hand, artificial intelligence and machine learning models have shown remarkable performance in automated evaluation of medical images \cite{hosny2018artificial}\cite{suzuki2017overview}\cite{shen2017deep}. In this situation, hospitals are increasingly drawn towards adopting computer-aided detection (CAD) technologies for processing scans. These technologies show considerable promise in improving diagnostic accuracy, reducing reporting time, and boosting radiologist productivity.

Supervised machine learning, the most common form of machine learning, works in two phases. In the first phase, the algorithm implemented as a software reads a training dataset consisting of images along with their corresponding labels. It processes this data, extracts patterns from it, and learns a function that maps an input image to its corresponding label. The learned mapping function along with the extracted patterns are mathematically represented in the form of the trained model. This is called the \textit{training phase}. In the second phase, called \textit{inference phase}, the trained model is used to read input images and make predictions. Artificial neural networks are a class of machine learning algorithms; artificial neural networks with many layers are called deep neural networks. In literature, the terms deep learning, artificial intelligence, and artificial neural networks tend to be used interchangeably. We use \textit{machine learning} in this paper to broadly refer to all the above terms in addition to conventional machine learning algorithms such as linear regression, support vector machines, decision trees, random forests, etc.

Developing machine learning models for radiology involves many challenges. High quality training data is vital for good model performance but is difficult to obtain. The available data may lack volume or diversity. It may be scattered across multiple hospitals. Even if image data is available, it may not be labelled. Radiology scans suffer from a high degree of inter-reader variability where two or more radiologists label the data inconsistently; this may lead to noise or uncertainty in the ground truth labels. The distribution of target classes may be heavily skewed, especially for rare pathologies. This imbalance in class representation is often accompanied by unequal misclassification costs across classes. Care must be taken when dealing with imbalanced datasets, and this sometimes requires using special performance measures. A model that works well on data from one hospital may perform poorly on data from a different hospital. Similarly, a model deployed in practice at a hospital may experience a gradual decay in performance at the same hospital. Machine learning models have been shown to be vulnerable to malicious exploits and attacks. To support adoption by radiologists, the deployed models should be able to explain their decisions, and they should not discriminate patients on the basis of gender, ethnicity, age, income, etc.

This paper follows a simple structure. In Section 2, we enumerate the key considerations that machine learning researchers should acknowledge and address. For each consideration, we describe common challenges and their significance before suggesting solutions to overcome them. In Section 3, we conclude by discussing other overarching limitations that hinder the adoption of machine learning in clinical radiology practice.

\section{Key Considerations}

\subsection{Insufficient Training Data}
Machine learning models are data hungry, and their performance depends heavily on the characteristics of the data used to train them \cite{foody1995effect}. Training set size has a direct and significant effect on the performance of the models. On the other hand, the heterogeneity and diversity of the training data influences the ability of the models to generalize to unseen data sources. To develop robust machine learning models, researchers need access to large medical datasets that adequately represent data diversity in terms of population features such as age, gender, ethnicity, medical conditions etc. and imaging features such as equipment manufacturers, image capture settings, patient posture etc. Most available datasets in medical imaging do not meet these requirements. Since many critical conditions have low rates of occurrence, very little data is available for them. Machine learning models trained using this scanty data to diagnose rare conditions fail to perform well in practice, even if they demonstrated good performance in retrospective evaluations.

Several methods have been proposed for dealing with insufficient data for training models. Data augmentation techniques including geometric transformations and color-space transformations can enhance the quantity and variety of the training data \cite{shorten2019survey}. Generative Adversarial Networks have shown success in generating synthetic images for rare pathologies which can further be used for model training \cite{yi2019generative}. Although these techniques allow models to be trained on scarce data by artificially increasing the variation in the dataset, they cannot serve as a substitute for high quality data.

\subsection{Decentralized Datasets}
Many medical datasets are naturally distributed across multiple storage devices connected to networks owned by different institutions. In traditional machine learning settings, these datasets need to be consolidated into a single repository before training the models. Moving large volumes of data across networks poses several logistical and legal challenges. Government policies such as General Data Protection Regulation (GDPR) \cite{voigt2017eu}, Health Insurance Portability and Accountability Act (HIPAA) \cite{annas2003hipaa}, Singapore Personal Data Protection Act \cite{chik2013singapore}, etc. also stipulate restrictions on sharing and movement of data across national borders.

Privacy preserving distributed learning techniques such as federated learning \cite{mcmahan2017communication} and split learning \cite{vepakomma2018split} enable machine learning models to train on decentralized datasets at multiple client sites without moving the data and compromising privacy. Implementing these techniques, however, entails additional overheads which may render the exercise unfeasible. These overheads include the high cost of developing software that supports these technologies, high network communication bandwidth required, the orchestration effort in deploying it at multiple sites, and the possibly reduced performance of the predictive models \cite{gawali2020comparison}. Federated learning generates a global shared model for all clients leading to situations where, for some clients, the local models trained on their private data perform better than the global shared model. In such situations, additional personalization techniques may be required to fine-tune the global model individually for each client \cite{kulkarni2020survey}.

\subsection{High Cost of Annotations}
Supervised machine learning requires radiology images to be annotated before they can be used to train the model. Image-level annotations classify each image into one or more classes, while region-level annotations highlight regions within an image and classify each region into one or more classes. Since the predictive performance of the model is directly influenced by the quality of the annotations, it is imperative that the data is annotated by qualified radiologists or medical practitioners. This makes the process of annotations exorbitantly expensive in many cases.

Several efforts have been made to use natural language processing (NLP) techniques to automatically annotate images by extracting labels from radiology text reports \cite{zech2018natural}\cite{smit2020chexbert}\cite{pons2016natural}. Semi-supervised approaches can be used when a small amount of labelled data is available along with a larger amount of unlabelled data \cite{cheplygina2019not}\cite{feyjie2020semi}. Since manual annotations are expensive, AI-based automated image annotation techniques can be considered \cite{cheng2018survey}.

\subsection{Ambiguous Ground Truth}
Since hospital datasets usually contain images accompanied by their text reports, many projects are kickstarted by using NLP techniques to automatically annotate the images using the reports. Radiology reports, however, vary widely in their comprehensiveness, style, language, and format \cite{brady2018radiology}. Even if state-of-art NLP manages to accurately extract all the findings from the text report, the report itself may not mention all findings. Olatunji et al. \cite{olatunji2019caveats} show that there exists a large discrepancy between what radiologists see in an image and what they mention in the report; reporting radiologists usually document only those findings that are relevant to the immediate clinical context and are likely to miss reporting non-actionable or borderline findings.

Radiology images suffer from significant inter-reader variability where two or more experts may disagree on the findings from a scan \cite{kerlikowske1998variability}\cite{rosenkrantz2018discrepancy}\cite{brouwer20123d}\cite{njeh2008tumor}\cite{moifo2015inter}\cite{irvin2019chexpert}. Sakurada et al. \cite{sakurada2012inter}, for instance, report low inter-reader kappa values ranging from 0.24 to 0.63 for assessment of different pathologies from chest radiographs. Annotation workflows in practice generally engage a single reader to assign ground truth labels to images. An improvement over this involves engaging multiple independent readers and considering their majority vote as the ground truth label. Single reader or majority vote approaches, however, may miss out labeling challenging but critical findings.

This risk can be mitigated by employing multiphase reviews \cite{armato2011lung} or expert adjudication \cite{majkowska2020chest} to create high-quality labels. Majkowska et al.\cite{majkowska2020chest} show that adjudication improved the consensus amongst radiologists to 96.8\% as compared to the 41.8\% after the first independent readings when assessing chest radiographs. Raykar et al. \cite{raykar2009supervised} propose a probabilistic approach to determine the \textit{hidden} ground truth from labels assigned by multiple radiologists and demonstrate that this method is superior to majority voting. In some clinical settings, radiology imaging is used for initial screening before conducting subsequent confirmatory tests. For example, chest X-ray scans may be used as a first-line test before subsequently conducting a CT scan, lab tests, or a biopsy. Data from these subsequent tests, if available, should be used to validate and correct the label assigned to the images from the screening test. In situations where human-labelled ground truth is noisy or ambiguous, developing a process to reduce variability and improve label quality may yield better models than attempts to improve model performance on the original labels by other means.

\subsection{Imbalance in Class Representation}
Class imbalance occurs when all label classes are not represented equally in the training dataset. This is a common situation when building binary classifiers for medical datasets where the number of \textit{normal} examples in which the target abnormality is absent is many times larger than the number of \textit{abnormal} examples in which it is present. Since machine learning models are usually trained by optimizing a loss function across all training examples, the trained models tend to favor the majority class over the minority one. Researchers have empirically evaluated the adverse effect of class imbalance on classification performance in several works: \cite{he2009learning}\cite{liu2011combining}\cite{kim2018impact}\cite{chen2017assessing}\cite{chawla2009data}\cite{luque2019impact}.

Class imbalance can be handled at the data level or the algorithmic level. Resampling strategies can be used to address imbalance in the training data by either undersampling the majority classes or oversampling the minority classes. Many comparative evaluations of these approaches exist, sometimes with contradicting conclusions. Drummond et al. \cite{drummond2003c4}, for instance, argue that undersampling works better than oversampling, while Batista et al. \cite{batista2004study} report superior performance using oversampling. However, we caution the reader against hasty generalizations and note that the aforementioned comparisons are highly dependent on the dataset, the machine learning algorithm, the sampling technique used, and the parameters of the experiments. Chawla et al. \cite{chawla2002smote} propose SMOTE (Synthetic Minority Over-sampling Technique), a technique to generate synthetic examples to balance the dataset, and show that the combination of SMOTE and undersampling performs better than plain undersampling or oversampling. Similarly, oversampling can also be performed using geometric augmentations, color-space augmentations, or using generative models to produce synthetic images. The imbalance in number of examples can also be addressed at the algorithmic level using methods such as one-class classification, outlier/anomaly detection, regularized ensembles, custom loss functions, etc. \cite{he2009learning}\cite{estabrooks2004multiple}\cite{yuan2018regularized}\cite{wei2018anomaly}\cite{ruff2018deep}.

\subsection{Asymmetric Misclassification Costs}
Standard machine learning settings assume that all misclassifications between classes are equal and incur the same penalty. This assumption does not hold true for many medical imaging problems. For example, the cost of classifying a \textit{normal} scan as \textit{abnormal} may be very different from the cost of classifying an \textit{abnormal} one as \textit{normal}.

This asymmetrical nature of the classification problem can be handled either at the time of deployment or during development. The trained model can be tuned for higher sensitivity or specificity as per the requirements at deployment time. Alternatively, the variation in misclassification penalties can be represented as a cost matrix, where each element $C(i,j)$ represents the penalty of misclassifying an example of class $i$ as class $j$. The model can then be trained by minimizing the overall cost as defined by this asymmetrical loss function. For more details, we refer the reader to literature on cost-sensitive learning \cite{elkan2001foundations}\cite{sun2007cost}\cite{he2009learning}.

\subsection{Relevant Performance Measures}
Machine learning researchers and practitioners tend to ignore the question of how model performance should be evaluated in cases of imbalanced datasets and asymmetric misclassification costs. Most binary classification models produce a continuous-valued output score. This score is converted into discrete binary labels using a cut-off threshold. Due to its simplicity, it is tempting to use accuracy - defined as percentage of predictions that are correct - as a measure of performance. In cases of imbalanced datasets, however, accuracy is ineffective and provides an incomplete and often misleading picture of the ability of the classifier to discriminate between the two classes \cite{maloof2003learning}\cite{joshi2001evaluating}.

Using two or more measures such as sensitivity, specificity, precision etc. provides a better picture of the discriminative performance of a classifier. These measures, however, are dependent on the cut-off threshold mentioned above. Furthermore, the decision on setting the threshold is often guided not by technology but by business or domain concerns. Comparing two models by considering multiple performance measures across different operating thresholds is challenging. The ROC curve, on the other hand, captures the model performance at all threshold operating points. The area under the ROC curve (AUROC) thus serves as a single numerical score that represents the performance of the model across all operating threshold points. This has made AUROC a metric of choice for reporting classification performance of machine learning models. Unfortunately, AUROC too can be deceptive when dealing with imbalanced datasets and may provide an overly optimistic view of performance \cite{he2009learning}. The precision-recall curve (PRC) and the area under it (AUPRC) are more suited to describe classification performance when datasets are imbalanced \cite{davis2006relationship}\cite{saito2015precision}. Drummond and Holte propose cost curves that describe classification performance over asymmetric misclassification costs and class distributions \cite{drummond2004roc}\cite{drummond2006cost}.

\begin{table}[hbt!]
\begin{center}
\begin{tabular}{ |c | c | c | c | }
\hline
 & Predicted as Negative & Predicted as Positive & Total\\ 
\hline
Actual Negative & 80 & 10 & 90\\
Actual Positive & 5 & 5 & 10\\
\hline
Total & 85 & 15 & 100\\
\hline
\end{tabular}
\caption{Example that illustrates how accuracy can be misleading in case of imbalanced datasets. In the above confusion matrix, out of 100 test examples, 90 are negative and 10 are positive. The classifier predicts 85 of them as negative and 15 as positive. This gives a high accuracy of 0.85 and a high specificity of 0.89. However, the complete picture is seen when we consider the low sensitivity of 0.50 and precision of 0.33.}
\end{center}
\end{table}

\subsection{Generalization of Models to Unseen Datasets}
Machine learning models are routinely evaluated on a hold-out set taken from the same source as the training set. The available data is divided into two parts. One part is used for training and validating the models. The second part, called the test set or hold-out set, is used to estimate the final performance of the trained model when deployed. The underlying premise is that the data used to train the model is representative of the data the model will encounter during clinical use. This assumption is often violated in practice, and this makes the performance on the hold-out set an unreliable indicator of the future performance in clinical deployment.

Poor generalization of models to diverse patient groups is one of the biggest hurdles for adoption of artificial intelligence and machine learning in healthcare. One reason for poor generalization is the difference in image characteristics between images from the training sites and those from the deployment site. This variation, also known as dataset shift, can occur due to differences in hospital procedures, equipment manufacturers, image acquisition parameters, disease manifestations, patient populations, etc. Due to dataset shift, models trained using data from one hospital may perform poorly on data from another hospital \cite{rajpurkar2020chexpedition}. We note here that this inability to generalize to datasets from an unseen origin is different from the problem of overfitting, where the model shows poor performance even on test sets from the same origin. Learning irrelevant confounders instead of relevant features is another reason why models fail to generalize to data from unseen origins. Machine learning models are notorious for exploiting confounders in the training data. For example, Zech et al. \cite{zech2018variable} show that a pneumonia classification model trained on data from two hospitals learned to leverage the difference between prevalence rates at the two hospitals instead of the relevant visual features. 

Data augmentation can improve model generalization by increasing the variation in the training set. Image processing techniques including standardization, normalization, reorientation, registration, histogram matching, etc. can be used to harmonize the images sourced from different origins and remove domain bias. Glocker et al. \cite{glocker2019machine} show, however, that even with a state-of-art image pre-processing pipeline, these techniques for harmonization were unable to remove scanner-specific bias, and machine learning models were easily able to discriminate between the different origins of the data.

Domain adaptation techniques can be used to fine-tune models to a new target domain by narrowing the gap between the source and target domains in a domain-invariant feature space \cite{ben2010theory}\cite{wang2018deep}\cite{ganin2016domain}\cite{long2016unsupervised}\cite{tzeng2017adversarial}. On the other hand, domain generalization techniques attempt to train models that are sensitive only to features relevant to the classification task but insensitive to confounding features that differentiate between the domains \cite{dou2019domain}\cite{bousmalis2016domain}\cite{li2018domain}\cite{muandet2013domain}\cite{volpi2018generalizing}\cite{peng2019domain}.

\subsection{Model Decay}
Model decay refers to the phenomenon where the performance of a deployed machine learning model deteriorates over time. Supervised machine learning algorithms extract patterns from the training data to learn a mapping between independent input variables and a dependent target variable. This process involves making an implicit assumption that the data encountered in deployment will be stationary and will not change over time; this assumption is often violated in practice due to changes in hospital workflows, imaging equipment, patient groups, evolving adoption of AI solutions, etc.

Model decay occurs due to changes in the underlying data. These changes can be broadly classified into three types: (1) \textit{Covariate shift} occurs when there are changes in the distribution of the independent input variables (e.g.: the average age of the population increases over time); (2) \textit{Prior Probability Shift} occurs when there are changes in the distribution of the dependent target variables (e.g.: the prevalence of a particular disease in the target population may change due to seasonality or an epidemic); and (3) \textit{Concept drift} occurs when there are changes in the relationship between the independent and the dependent variables (e.g.: changes in a hospital's diagnostic protocols or a radiologist's interpretation regarding which visual manifestations should or should not be considered indicative of a pathology). These changes can be sudden, gradual, or cyclic.

Detecting model decay requires continuous monitoring of the deployment-time performance against a human-labelled subsample of the data. If the performance drops below a predetermined threshold, an alarm is triggered, and the model is retrained or fine-tuned on the most recent data. This retraining can also be conducted periodically as a routine maintenance activity. For more details including theoretical frameworks for understanding model decay or practical solutions, we refer the reader to these reviews: \cite{widmer1996learning}\cite{vzliobaite2010learning}\cite{wang2018systematic}\cite{gama2014survey}\cite{vzliobaite2016overview}.

\subsection{Adversarial Attacks}
An adversarial example is constructed by deliberately injecting perturbations in the original image to fool the model into misclassifying it \cite{szegedy2013intriguing}. Machine learning models are susceptible to manipulation using such adversarial examples \cite{goodfellow2014explaining}\cite{moosavi2016deepfool}. Data poisoning attacks \cite{steinhardt2017certified} introduce adversarial examples in the training data to manipulate the diagnosis of the model being developed. Evasion attacks \cite{biggio2013evasion}, on the other hand, use adversarial examples to influence predictions during deployment. Healthcare is a huge economy, and many decisions regarding diagnosis, reimbursements, and insurance may be governed or assisted by algorithms in the near future. Hence, the discovery of these vulnerabilities has raised pressing concerns regarding the safety and usability of machine learning models in clinical practice.

Qayyum et al. \cite{qayyum2020secure} provide a detailed taxonomy of defensive techniques against adversarial attacks by grouping them into three broad categories:  (1) reconstructing the training or testing data to make it more difficult to manipulate \cite{goodfellow2014explaining}\cite{huang2015learning}\cite{gu2014towards}\cite{xu2017feature}\cite{gao2017deepcloak}; (2) modifying the model to make it more resilient to adversarial examples \cite{papernot2016distillation}\cite{katz2017reluplex}\cite{ross2017improving}\cite{bradshaw2017adversarial}\cite{nguyen2018learning}; (3) using auxiliary models or ensembles to detect and neutralize adversarial examples \cite{metzen2017detecting}\cite{lu2017safetynet}\cite{gopinath2017deepsafe}\cite{tramer2017ensemble}\cite{song2017pixeldefend}. Adversarial attacks and their countermeasures is an evolving research area, and there are excellent reviews for the same \cite{finlayson2019adversarial}\cite{chakraborty2018adversarial}\cite{akhtar2018threat}.

\subsection{Explainability}
The power of neural networks to uncover hidden relationships between variables and use them to make predictions is tempered by one disadvantage: the exact process the neural network uses to arrive at a decision is opaque to humans. This is why neural networks are sometimes called black boxes whose inner workings we cannot observe. To what extent can we delegate decision making to machines while we remain unaware of how the machine arrives at a decision is a key question that stands in the way of adoption of algorithms in many industries including autonomous vehicles, law, finance etc. 

Algorithmic explainability is especially important in medicine, where the stakes are high and the field is prone to litigation. In the context of radiology, explainability can be improved by using localization models that highlight the region of interest within the scan that is suspected to contain the abnormality instead of classification models that only indicate presence or absence of the abnormality. Developing localization models, however, also requires the training data to have region-based annotations in the form of bounding boxes or free-form masks. Where region-based annotations are not available, saliency maps \cite{selvaraju2017grad} and explainability frameworks \cite{ribeiro2016should} can be used to identify a region within the image that most contributed to a particular decision. Another way to improve the user's trust in the models is to predict a confidence score in addition to the prediction. For example, instead of merely stating the prediction ``Probability of Tuberculosis: 75\%'', the system should also state the model's confidence ``Probability of Tuberculosis: 75\%, Confidence in this prediction: Low''. Deployment settings where predictive models are used to autonomously make decisions demand more stringent conditions of explainability than settings where the models are used to guide humans who make the final decisions.

There have been calls to limit the use of artificial intelligence and machine learning only to rule-based systems in fields where algorithmic decisions affects human lives \cite{campolo2017ai}. These systems are transparent and can trace the relationship between the input and the output as a sequence of rules that humans can understand. We find two problems with this approach. First, one of the chief advantages of using neural networks is that they can model complex relationships that \textit{humans cannot understand}; and this is precisely what makes them so effective. Second, making decision systems transparent and explainable also makes them vulnerable to malicious attacks. A transparent rule-based method to make decisions can be `hacked', `gamed', or exploited more easily than a black box system.

\subsection{Fairness and Bias}
Algorithmic systems play a key role in guiding decisions that impact the delivery of healthcare to patients. It is desirable, therefore, that these systems are free of societal biases, and their decisions are fair and equitable. Unfortunately, many existing datasets reflect the biases of the societies which they represent, and it is difficult to detect and remove bias that is inherent in the training data. Obermeyer et al. \cite{obermeyer2019dissecting} show, for example, that a widely used algorithmic system exhibited racial bias against Black patients that reduced the number of Black patients eligible for extra care by more than half.

In principle, a predictive model is considered fair if it does not discriminate patients on the basis of sensitive variables such as gender, ethnicity, disability, income, etc. Translating this seemingly simple principle into practice, however, is a challenging affair. Researchers have developed numerous mathematical definitions for fairness and techniques to implement them \cite{verma2018fairness}. One technique, for example, is to exclude sensitive variables from the input when training the model. Another technique is to tune the model so that it demonstrates the same level of performance as measured by sensitivity, specificity, etc. across all groups defined by the sensitive variables. Corbett-Devies and Goel \cite{corbett2018measure} show that these techniques, although appealing, suffer from significant statistical limitations, and they may adversely affect the same groups they were designed to protect. Pleiss et al. \cite{pleiss2017fairness} show how different definitions of fairness can be mutually incompatible, and a model designed to comply with one definition may violate another equally-valid definition.

Algorithmic bias and fairness is an evolving field of research that lies at the intersection of machine learning, public policy, law, and ethics. We believe that fairness is not inherently a technological problem but a societal one. Coercing technology to solve it can lead to automated systems that tick the right boxes for some arbitrary definition of fairness but eventually end up worsening social inequality and discrimination behind a veneer of technical neutrality \cite{benjamin2019assessing}.

\subsection{Clinical Validation}
A comprehensive evaluation to assess the predictive performance and clinical utility of a model must be conducted before it can be deployed in clinical practice. When a model is evaluated on a hold-out set collected from the same sources that the training data is collected from, the evaluation is called an \textit{internal} validation. When a dataset from an \textit{unseen} source is used to evaluate the model, the evaluation is called an \textit{external} validation. As described above in section 2.8, lack of generalization to unseen data sources is one of the biggest challenges in adoption of machine learning in practice. Despite this, only a fraction of published studies report results of an external validation \cite{kim2019design}. Mahajan et al. \cite{mahajan2020algorithmic} present examples to advocate the case for independent external validation of models before deployment and describe a framework for the same. Park et al. \cite{park2018methodologic} propose a methodology with a checklist for evaluating clinical performance of models. The TRIPOD statement \cite{collins2015transparent} provides guidelines for transparent reporting of development and validation of prediction models for prognosis and diagnosis models. While retrospective evaluations allow machine learning developers to test their models on large and diverse datasets, prospective evaluations allow testing in real-world environments; both types of evaluations are equally important and should be meticulously carried out before full-scale adoption.

\section{Conclusion}
We identify key challenges that researchers face in developing accurate, robust, and usable machine learning models that can create value in clinical radiology practive. These challenges and the techniques to overcome them have been discussed previously in a piecemeal manner in prior research literature. In this paper, we re-examine them in the context of medical imaging. By compiling them in the form of a laundry list, we hope to make this research more readily accessible.

Hospital workflows and practices vary widely from one hospital to another, even within the same geography. This increases the difficulty in seamlessly integrating predictive models into the hospital workflow. The non-uniformity in workflows also raises the question whether the reported performance of a model is reproducible under a different clinical context. This is on-going research and satisfactory solutions are yet to be found.

The ultimate objective of diagnostic machine learning models is to improve patient outcomes. However, improvement in diagnostic performance does not, by itself, cause an improvement in patient outcomes \cite{park2018methodologic}. Radiological diagnosis is only one of many steps that eventually leads to treatment. A computerized diagnostic system, therefore, must be appropriately placed in the workflow. How the system presents the results to the reporting radiologist and what action the radiologist takes on receiving them are important factors that influence how useful the system will be in practice.

Medical imaging is a broad and complex field that encompasses numerous imaging modalities, pathological conditions, and diagnostic protocols. On the other hand, machine learning is an active area of research with thousands of new techniques published every year. The combined diversity of both fields along with non-uniform hospital practices, regulatory restrictions on data sharing, and lack of standardized reporting of results makes it difficult to clearly assess the role and potential of machine learning applications in medical imaging. We believe machine learning has great potential in improving diagnostic accuracy, lowering reporting times, reducing radiologist workloads, and ultimately improving delivery of healthcare. To realize this potential, however, a concerted across-the-board effort will be required from physicians, radiologists, patients, hospital administrators, data scientists, software developers, and other stakeholders.

\bibliographystyle{ieeetr}
\nocite{*}
\bibliography{bibliography}
\end{document}